# Memory via Temporal Delays in weightless Spiking Neural Network

Hananel Hazan[1], Simon Caby, Christopher Earl[2], Hava Siegelmann[2], Michael Levin[1]

1 Dept. of Biology and Allen Discovery Center at Tufts University, Medford, MA, 02155-4243, USA
2 University of Massachusetts, Amherst, MA, USA

*Abstract*

A common view in the neuroscience community is that memory is encoded in the connection strength between neurons. This perception led artificial neural network models to focus on connection weights as the key variables to modulate learning. In this paper, we present a prototype for weightless spiking neural networks that can perform a simple classification task. The memory in this network is stored in the timing between neurons, rather than the strength of the connection, and is trained using a Hebbian Spike Timing Dependent Plasticity (STDP), which modulates the delays of the connection.

## INTRODUCTION

The role of synaptic plasticity in learning and memory is an open question in neuroscience. The current understanding of synaptic modification dates to Hebb's postulate [1], which states that if cell *A* persistently excites cell *B*, cell *B* will be more easily excitable by *A* (colloquially, *neurons that fire together, wire together*). This postulate and its notion of past activity remodeling synapses is the basis for modern research done in neural computation. This perception may have also influenced modern computer science to use numerically weighted connections during learning, in addition to weighted connections being a mathematical convenience for modulation.

An alternative mechanism for manipulating information and storing memory is regulating the speed of spike propagations between neurons [2], [3]. Myelin, although only present in vertebrates, is a crucial element of neurons in the central nervous system. Similarly in non-vertebrates, the thickness of the axon can change the speed of spike propagations (other analogous mechanisms to the functionality of myelin in invertebrate neurons may also be found). Myelin is a fatty tissue that surrounds nerve axons and acts as electrical insulation. This insulation allows the electrical signals (spikes) to travel further and with less degradation. More importantly, myelinated axons can propagate spikes much faster and with less energy consumption [4], [5] compared to unmyelinated axons. Experimental findings on the nervous system have shown that delay times can vary between organisms. For example, Sprague-Dawley rats have delays of 1 to 17 ms [6], rabbits 1 to 32 ms [7] and cats 1 to 30 ms [8]. Furthermore, the amount of myelin around axons is not constant and can be increased or decreased during myelination and demyelination [8]–[10] respectively.

In this paper, we set out to explore different architectures for plasticity in computational learning models, with a focus on biological plausibility. We emphasize biological plausibility because we believe the empirical success of biological networks can provide insights to developing high performance learning models, as well as further our understanding of neural computation. We present a proof of concept, that *timing* of spike transmission alone can be used for computations in biology. In our simple neural network, which we call a *weightless spiking neural network (WSNN)*, learning occurs not with weight adaptation but with spike propagation times. We show that our weightless networks can be trained on the MNIST dataset in an unsupervised manner, resulting in comparable performances with unsupervised weight-based models. The purpose of this paper is to suggest an alternative to weight-based networks using delays as a means to modulate synchrony and spike intensity.

## Previous work

Neurons in biology are believed to operate in an asynchronous manner, i.e. they can process and generate signals at any point in time, without the need for a clock which dictates their activity. This has led to the popular belief that the timing between spikes may be used as an additional dimension for information encoding, memory storage, memory capacity, and a possible mechanism for decision-making.

In decision making, the timing between spikes can be used as a faster mechanism to react to external stimuli, a quality that can give the organism an evolutionary advantage for survival. Several papers propose to use the scheme of Time To First Spike (TTFS) [11], [12] for classification tasks while providing the minimal possible information to the network. This in turn makes the network significantly faster as there is less information to process.

Other approaches use the frequency of communication between neurons as clues for storing information and performing computations using oscillations with networks based on timing and delays. Examples for such a network can be found in Izhikevich [13],[14] that present the polychronization network model. The model was randomly initialized and exhibited self-organization into groups of neuron clusters with different firing patterns and shifts between different firing oscillations. The author of [13] suggests that the number of different groups that generate unique patterns exceeds the number of neurons in the network, evidence for the high memory capacity of the system. Wright et al [15] demonstrated an algorithm that modifies the mean and variance of postsynaptic spikes during training. Using this method, the authors were able to recognize a temporal sequence of spike trains in both supervised and unsupervised learning schemes.

Lastly, the authors Zhang et al [16] proposed a supervised learning rule that updates both the weights and delays of a synaptic connection. Using this learning rule, the authors trained a network on the TIDIGITS corpus, and concluded that the success of their model demonstrates that a combination of weights and delays can surpass the standard weighted network.

The above works show the latent potential of using delays in the connections. Our approach in this paper is taking another step forward by presenting an unsupervised learning rule that also works on static data and operates on the level of the individual neuron and their weightless connection.

# METHOD

The networks in this paper are built with BindsNET [17], a flexible and open source framework for experimenting with spiking neuron architecture.

We propose a biology-inspired myelination process to modify the axonal delays of a 2-layered, feed-forward, Spiking Neural Network (SNN) with no synaptic weights. The network uses a delay-encoded input, and an output layer which utilizes a TTFS scheme. Competition in the output layer is enforced in two dimensions:
1. A sliding threshold that tunes the firing activity of each neuron
2. By a Winner-Take-All mechanism (WTA) where the first spike in the output layer prevents all other neurons in the output layer from firing. This is simulated as a full lateral inhibitory connection similar to [18], [19].

A linear decoder (readout) analyzes the outputs of each sample, and statistically assigns output neurons to input MNIST digits. Note that the linear decoder applied directly to the MNIST dataset reaches an accuracy of 62% by itself. For the sake of computational costs, simulations were stopped after the first emitted spike(s) on the output layer.

## Network architecture

Output neurons in the network are Leaky Integrate and Fire neurons (LIF) similar to [18], [19] with an adaptive firing threshold. A unique feature of the spiking neurons is their ability to process spatio-temporal signals and transmit them in spike form. These neurons can take advantage of their multidimensional capabilities by encoding information propagated through the network in the form of precise spiking times, ordered spiking sequences (bursts), and spike frequency.

Leaky Integrate and Fire Neuron (LIF) Equations:

$$\frac{dv}{dt} = (v_{rest} - v\ ) + g_e(E_{exc} - v\ ) + g_i(E_{inh} - v\ )$$

One common issue with SNN's is the oversaturation of signals. For example, if the firing thresholds for neurons in a network are too low, they will become highly sensitive to incoming input. This will result in excessive spiking from neurons in the network, subsequently making information gain very difficult. This is comparable to a seizure seen in biological networks. To combat this, the LIF neurons used in the following experiments utilize a threshold adaptation mechanism which adjusts the sensitivity of the neuron to the quantity of incoming spikes. To do this, we utilized two variables, $\theta_0$ and $\theta_1$, to modulate the value of the membrane threshold $T_i$ at timestep $i$:

$$T_i = \theta_0 + \theta_i$$

In this formula, $\theta_0$ represents a constant base value for the membrane threshold which does not change during runtime. The variable $\theta_i$ acts as a modulator to the membrane threshold, and adapts according to behaviors observed at each timestep $i$. As more outgoing spikes are produced by the neuron, $\theta_i$ increases in value and thus increases the total value of the membrane threshold. The value of $\theta_i$ is subject to a constant decay at each timestep, allowing the neuron to also adapt to scenarios where spiking is sparse. This threshold adaptation helps neurons to maximize information gain and encode inputs in a more meaningful way, by avoiding signal saturation and over-sensitivity to input intensity changes. Moreover, we hypothesize that adaptive thresholds add a form of temporal memory to the neuron, since certain repeated or recognizable patterns will trigger varying amounts of activity in different regions of the network. This is an ability that we attempt to capitalize on in the learning algorithm. The results of our experiments and those mentioned in the background section supports this claim.

$$\frac{d\theta_i}{dt} = -decay * \theta_i + s_+$$

To optimize hyper-parameter searches, we increased the competition between neurons by forcing at least one of them to have a theta value of zero at all times:

$$\theta_{i(t+1)} = \theta_{i(t)} - min(\theta)$$

The learning rule used was a homeostasis process between sliding threshold and Hebbian STDP (as a result from the WTA). Together, they create a competition mechanism that tunes the spike delays.

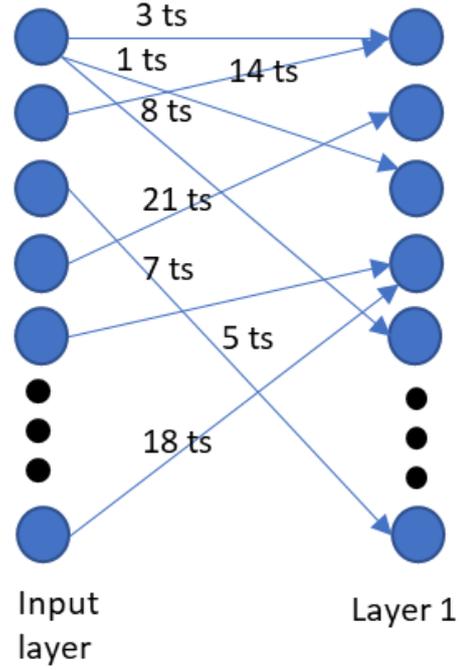

*Figure 1 - illustration of Network schematic between the input layer to the first layer. ts=time step.*

## Network Parameters & Initialization

The WSNN networks used in the following experiments are composed of two primary parameters:

1. **Synaptic Delays**: Spikes travel in a myelinated axon much easier and faster compared to an unmyelinated axon [5], [20]. In our network delays are represented by integer values measuring the number of time steps (milliseconds) to delay an outgoing signal by. Initial values were sampled from a uniform distribution set to a range of [0, 32], and rounded to integer values.

2. **Firing Thresholds**: Thresholds determine the required membrane potential (volts) of an LIF neuron to generate a spike. Recall that LIF neurons gain charge through excitatory connections and lose charge through inhibitory connections and membrane leakage. The constant portion of the threshold $\theta_0$ is initialized to a value of -52, and $\theta_i$ to a value randomly sampled from a uniform distribution in the range [0, 0.1].

## Inter-neurons delay plasticity

is a popular, biologically inspired learning rule which has been successfully used to train Spiking Neural Networks (SNN) in both supervised and unsupervised paradigms. As opposed to traditional Artificial Neural Networks (ANN), signals in SNN's are propagated in the form of spikes emitted at precise times. Inspired by mechanisms observed in biological networks, STDP (Hebbian or anti-Hebbian) has been widely adopted to train the relative weights of SNN's.

Currently, most SNN designs use a static delay time between neuron layers (usually the length of one-time step used in the simulation), if any at all. For our WSNN's, we've chosen to use an emulation of myelin insulation which allows for dynamic adjustments of delay times. Because myelination can modulate the timing of spikes, using it in conjunction with STDP can tune neurons to spike in synchrony. The synchronized firing of neurons has shown to be a critical mechanism for learning [21]–[24] and decision making [25] in the human brain, and has also proven to be effective in previous works with weighted SNN's. As described previously, weighted networks are reasonable replications of biological networks, but are still widely regarded as biologically implausible. Thus, we have chosen to attempt replacing such weights with time delays.

In mammalian brains, STDP is one of the known synaptic adaptation mechanisms, which dynamically modifies the properties of afferent neurons. In general, STDP is a learning process that follows the principles of Hebbian learning (see Intro.) through the utilization of long-term potentiation and long-term depression. In previous works with SNN's trained on STDP, changes in weights were primarily responsible for the adaptation of information gain through synchronization. Intuitively, if the synaptic connection between a source (A) and target neuron (B) sees a weight increase, then the probability of B firing as a result of activity from A increases (note that the inverse of this statement is also true, where depressed connections reduce the probability). A similar conjecture could be made for dynamic delays: if the synaptic connection between A and B sees a delay decrease, then the probability of A firing as a result of activity from B increases. For our experiments, we have chosen to adopt this interpretation of STDP to train our networks.

Like most neural networks in machine learning, Our WSNN models use several hyperparameter values which can be optimized for performance. For this process, we chose to use a PSO (Particle Swarm Optimizer) algorithm. Optimal hyper parameters values are reported in Table 2:

- **Additive Decay**: a small constant continuously added to the normalized synaptic delays.
- **Encoding Time**: the maximum delay (in ms) for encoding the input signal using TTFS.
- **Learning Rate**: the maximum magnitude of STDP changes to the delay connection depending on the timing of the stimulus.
- **Max. Synaptic Delays**: the longest possible inter synaptic delay.
- **Neuron Threshold**: the base neuron threshold value $\theta_0$.
- **Delay Norm**: the normalization value of the synaptic delays, ranging from 0 to 1.
- **Spike Intensity**: the voltage increase applied to a neuron for each of its incoming spikes. All spikes in the network produce the same spike intensity.
- **Theta Plus**: the constant voltage increase of a neuron's membrane threshold after an incoming signal is received. This constant voltage increase is represented by $\theta_i$ for neuron $i$.

The proposed modified Hebbian STDP rule applies to the transmission delay of the axons, according to the delay between afferent and efferent spikes $\Delta_t$:

$$dm_{ij}/dt = -\mu . e^{-\Delta_t/\tau}, if\ \Delta_t > 0$$

where $m_{ij}$ represents the transmission delay of the axon between cell $i$ and cell $j$, and $\mu$ is the learning rate.

Note that $\Delta_t$ is computed at the neuron level, between incoming spikes and outgoing spikes, regardless of the incoming spike transmission time.

The transmission delay (expressed in number of dt) of a synapse is evaluated as:

$$d_{ij} = \lfloor max\ delay * m_{ij} \rfloor$$

A delay buffer in each synapse simulates the variable transmission delay between neurons.

Similar to most STDP-based SNN, we apply a normalization mechanism to the transmission delay, reflecting the locally limited biological resources to construct the myelin sheath:

$$m(t+1)_{ij} = m_{norm} * \frac{m(t)_{ij}}{\sum m(t)_{ij}}$$

A slow constant increase of the transmission delay is also simulated, reflecting a natural degradation process of the myelin sheath (demyelination). This allows neurons to focus on the most common patterns, while less common patterns are slowly forgotten (known as catastrophic forgetting):

$$m_{ij(t+1)} = min(m_{ij(t)} + c, 1)$$

Figure 1 illustrates the full connection from input to output layer, with variable numbers of the time step for each connection.

### Input encoding

For our experiments, we used the MNIST dataset to test our networks. Preprocessing of samples consisted of normalizing image data to reduce imbalance between dark and bright images. As samples are passed into the WSNN, input neurons fire only one spike using a TTFS scheme [26] at a time proportional to the pixel's brightness/intensity. Brighter pixels will spike earlier, with black pixels emitting no spikes at all. The encoding time window is chosen in the range 20 to 32ms. The motivation for using TTFS is the reduced number of spikes, and reduced simulation time. [27] showed that models using TTFS-encoded images perform similarly to those using Poisson-encoded images (Figure 2), in network topologies similar to ours.

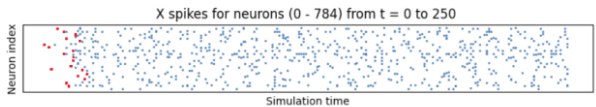

*Figure 2 - Poisson encoding of digit 8 and the red dots are the TTFS encoding.*

## RESULTS

Networks were evaluated using 1000, 2000, 3000, and 4000 neurons, and their performances are shown in Figure 5. Accuracies were reported after 1 epoch, as Further training led to negligible accuracy increases.

During training, the average number of simultaneous output spikes decreased from 20 to ~2.5, with correctly predicted digits leading to fewer simultaneous spikes (about 30% less). Additionally, time to the first output spike decreased, with correctly predicted digits leading to shorter output spiking times (about 6% shorter) as shown in Table 1 and Figure 3. The network schematic can be shown in Figure 1 and results can be found in Figure 4, Figure 5, Figure 7, and Table 2.

By using only Time To First Spike (TTFS) and delay-learning, our model takes less time (iterations) to compute and has less spikes to predict when compared with equivalent weight-based models using Poisson encoding [18] see Table 1.

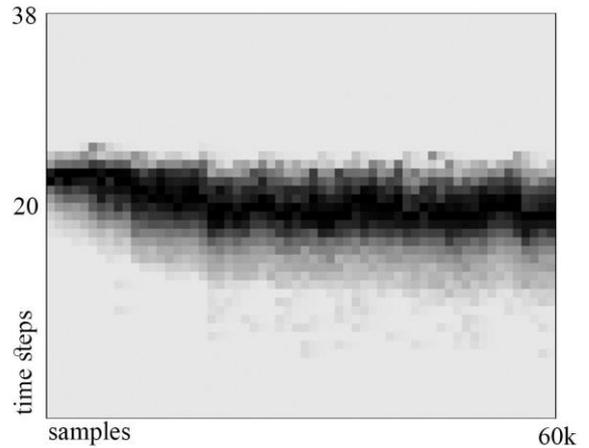

Figure 3: Shows the distribution of the 3000 neurons output spikes timings (X axis, in timesteps units) throughout one training epoch (Y axis, in number of samples). From the figure we can see the decrease of the average spike timing of the neurons towards a stable value in the learning process

|  | Our model | Weight-based (w Poisson) |
|---|---|---|
| time-steps | 10-25 ms | 250 ms |
| Total spikes | 80-100 | 500-1000 |

*Table 1: the table shows a typical number of spikes and the duration of one image presented to the model.*

| Network Size | Additive Decay | Encoding Time (ms) | Learning Rate | Max Synaptic Delays (ms) | Neuron Threshold (mV) | Delay Norm | Spike Intensity (mV) | Theta Plus (mV) |
|---|---|---|---|---|---|---|---|---|
| **1k** | 0.000069 | 31 | -0.0286 | 62 | -50 | 0.51 | 0.53 | 1.9 |
| **2k** | 0.000048 | 25 | -0.0474 | 40 | -58 | 0.64 | 0.31 | 1.4 |
| **3k** | 0.000045 | 20 | -0.0374 | 28 | -51 | 0.73 | 0.41 | 1.5 |
| **4k** | 0.000034 | 24 | -0.059 | 37 | -53 | 0.6 | 0.34 | 1.9 |

*Table 2 Parameters for the best performance network with 1K, 2K, 3K, 4K neurons*

Final delays are quantizable and can be encoded with only 5 bits (maximum synaptic delay < 32-time steps). The memory footprint for a 4000-neuron model with 5-bit encoded delay and 28x28 inputs is about 2 MB.

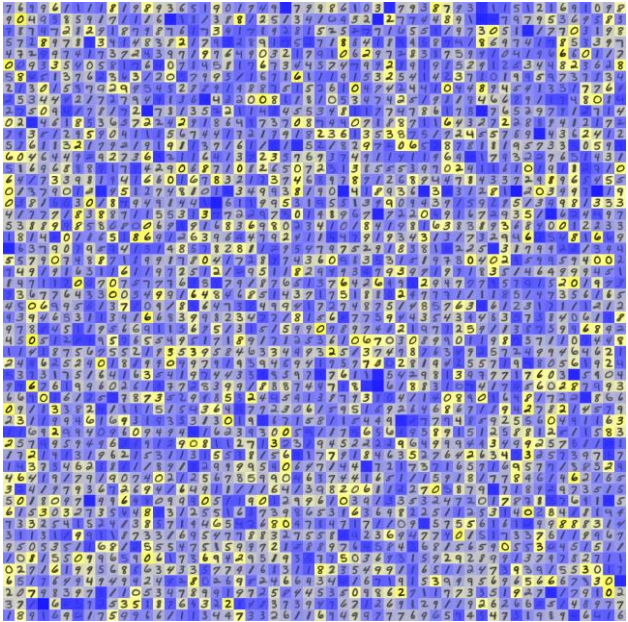

*Figure 4: Trained synaptic delays for 3k WTA network. The figure shows the connection delay values between the WTA layer to the input layer. Each neuron has 28x28 values, reorder into MNIST image shape. Darker pixels in the patterns show the shortest delays. Yellow patterns have the highest threshold offset $\theta_i$ values, blue ones have the lowest.*

Table 2 demonstrates an accuracy comparison with a similar architecture and learning scheme from [19], using a weights-based STDP model, trained for 1 epoch:

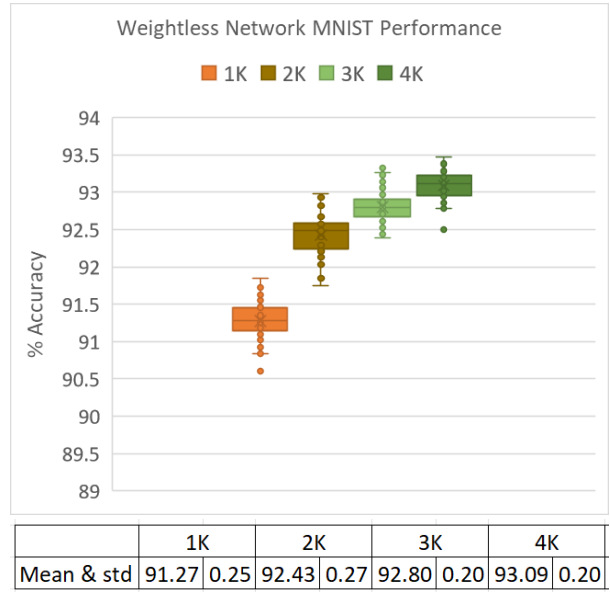

| | 1K | | 2K | | 3K | | 4K | |
|---|---|---|---|---|---|---|---|---|
| Mean & std | 91.27 | 0.25 | 92.43 | 0.27 | 92.80 | 0.20 | 93.09 | 0.20 |

*Figure 5: algorithm performances on variance network size (1K, 2K, 3K and 4K). The results presenter in the table are the average of 30 runs and their standard deviation*

## Limitations

Looking at the misclassified MNIST (see Figure 6) digits clearly shows the limitations of our model. With enough misplaced bright pixels, a samples input pattern can induce early firing of a wrong output neuron. Penalizing images for having excess white pixels was attempted but did not improve the model. We believe that this issue could be mitigated by adding another layer of WTA, or another layer that uses TTFS encoding scheme with inhibition. As seen in [28], using twin excitatory and inhibitory input layers can add the computational power to distinguish between similar input patterns, but will also require a more complex learning mechanism, ie. Hebbian STDP with two simultaneous input layers.

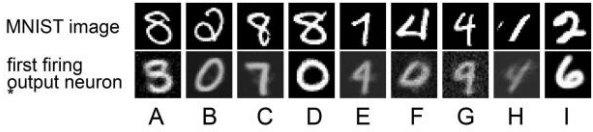

*Figure 6: examples of misclassification. The first line contains the MNIST image ground truth, and the second line the neuron classification depicted from the neuron delay values*

The confusion matrix (see Figure 7) highlights the need for an inhibition of the excess white pixels. The image depicting the digit "8" often triggers output neurons associated with digits "5", "1" and "3", whenever there are enough correctly placed white pixels.

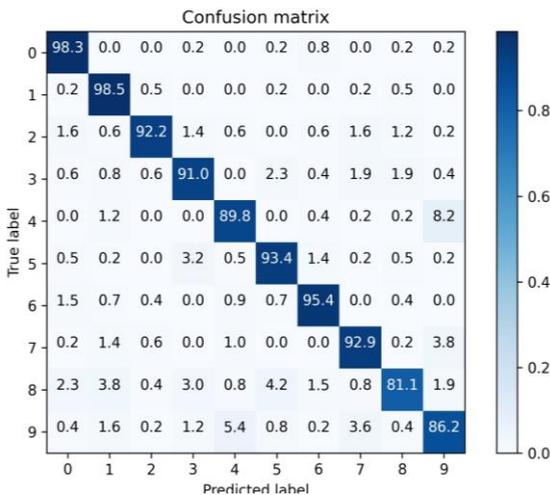

*Figure 7: Typical confusion matrix, 3k output neurons*

In classic weight based SNN models, the temporal dimension is discretized for the whole network and represented by the dt parameter. This discretization helps to compute the voltage activity of neurons in a more computationally efficient fashion (more efficient than solving the differentials in the *Network Architecture* section), but will also reduce the accuracy of spike times. Intuitively, a higher granularity will provide a closer replication of a continuous time interval, and thus a higher accuracy. Weights themselves are unaffected by the discretization of time because the weights as they only modify the magnitude of the output, not the timing; however, the same cannot be said for delays. Delays directly modify the outgoing times of spikes, and must be measured using multiples of dt. If the value for dt is too large and inaccurate, then the delays will suffer as well. As a result, we believe WSNN's have an inherent reduction in precision due to their dependence on delays.

In our model, the loss of precision due to time quantization was hard to estimate precisely, since every attempt with a different maximum TTFS encoding value for the delays needs a complete reconfiguration of the other parameters. This needs further examination.

The TTFS-encoded input may prove more beneficial in the case of a dataset containing a more evenly distributed set of pixel values, whereas black and white pixels comprise over 90% of all pixels in the MNIST dataset. Further research should include less contrasted inputs, such as CIFAR.

Another issue with this model is its sensitivity to hyper-parameter values. One potential issue is that using TTFS does not guarantee an output spike will occur. This can happen when firing thresholds for neurons are set too high, resulting in too little spiking activity to reach the output layer. This limits the perimeter of the valid parameters space, as we need at least one output spike for the model to make a prediction. The possible values for plasticity parameters (neuron's thresholds and delay-STDP learning rates) are therefore limited.

The drawback of this approach is that the current neuronal model solely relies on the time aspect of the first spike from the input and can miss some of the existing information in the static or timeless data. Moreover, the presented model uses a winner take all architecture [18], [19] to reduce the excessive spikes that are left in the system after the winner has been chosen. To alleviate the excessive spike in this model, IF (Integrate and Fire; no 'leaking') neurons can replace LIF with no loss in accuracy. This can be tested by setting the neuron's decay speed to infinity.

# DISCUSSION

We presented a proof of concept for a time-based model which uses only the timing of the spikes and the firing thresholds to perform classical classification tasks. To emphasize the fact that connections do not have numerical weights, we call our model Weightless Spiking Neural Networks (WSNN). These networks utilize a learning strategy that is consistent with observations in neuroscience that suggest spike timing conveys information for learning and decision making [2], [3]. The successful learning capability of our model suggests that the mechanism which changes the speed and the conductances of spike transmissions can also be part of learning and event encoding in neuronal models. Moreover, changing the spike transmission timing seems to be a natural continuation for known learning rules that rely on the timing and frequency of stimuli.

In the proposed WSNN, synaptic adaptations are done using the standard Hebbian STDP rule to adjust the communication timing between neurons. The result of the proposed learning scheme is that the information encoded in the delay connection can use both the rate of the spikes (rate code) and the timing of the spikes (TTFS) to convey and encode information. Additionally, this network uses learning rules which utilize both the temporal and spatial dimensions of SNN's, adjusting both the quantity (neuron adaptive threshold) and timing (synaptic delays) of the spikes. In comparison, prior models only dealt with the spatial aspect by using synaptic weights to influence only the total spike quantity.

Motivated to address the biological implausibility of weighted connections, the model presented in this paper shows that the manipulation of communication delays between neurons is a viable method for training SNN's and can be used in replacement of weights. This learning method also benefits from being compatible with Hebbian learning rules, such as STDP, which were originally designed to operate on a continuous-time interval. As such, rules like STDP were a natural and intuitive method of modulating delays.

Using delay-based connections can benefit threefold: shorter decision times, fewer computational resources, and reduced energy consumption. Since the network operates on TTFS, the decision can be reached significantly faster (see Table 1). The computational benefits of using delayed connections can be addressed in two ways: number of computational operations and memory utilization. The network uses only binary signals, meaning significantly fewer operations and resources are needed to store, train, and simulate the network. Altogether, a shorter run, fewer computations, and less resources makes this model ideal for low resource environments like those found in biology.

Addressing the time aspect of connections between neurons as a means to prioritize communication between cells can open new possibilities for learning. Moreover, addressing the need to adjust the time delays between neurons opens up new dimensions for training algorithms. Furthermore, most of the information around us is time-driven data where time is an integral dimension to the data, and itself contains information. Considering that traditional neuronal networks do not excel with dealing with the time dimension and since the use of Hebbian learning rules like STDP and others as a training algorithm in spiking neuronal networks didn't yield the performance that we hope. Seems reasonable to assume that more properties in a conjunction with the Hebbian rules are needed. Adding the time aspect to the connection has the potential to that edge.

Future work will focus on models with more layers and higher neuron complexity that benefit from the timing aspect of the delay connections. Furthermore, using time-driven data such as video or sound can use the true potential of the delay network that uses the time aspect in the information to make a decision.

# Acknowledgment

We would like to thank Peter Marathas for his help in the writing and programming.

The simulation done in this paper where run on two HPC's, the authors acknowledge the Tufts University High Performance Compute Cluster (https://it.tufts.edu/high-performance-computing) and the Extreme Science and Engineering Discovery Environment (XSEDE) Bridges GPU Artificial Intelligence at Pittsburgh Supercomputing Center through allocation CCR200032.

M.L. gratefully acknowledges support of The Elisabeth Giauque Trust and the Barton Family Foundation.